%% file: main.tex
\documentclass[10pt,twocolumn,letterpaper]{article}

 \usepackage[pagenumbers]{style} 
\usepackage{makecell}
\usepackage{wrapfig}
\input{preamble}

\definecolor{blue}{rgb}{0.21,0.49,0.74}
\usepackage[pagebackref,breaklinks,colorlinks,citecolor=blue]{hyperref}

\title{\vspace{-2.0cm}\ourmethod{}: Multiview Inpainting for Fast Editing of 3D Objects}

\author{Amir Barda\\
Tel Aviv University, Israel \\
{\tt\small amirbarda@mail.tau.ac.il}
\and
Matheus Gadelha\\
Adobe Research, USA\\
{\tt\small gadelha@adobe.com}
\and
Vladimir G. Kim\\
Adobe Research, USA\\
{\tt\small vokim@adobe.com}
\and
Noam Aigerman\\
Université de Montréal, Canada\\
{\tt\small noam.aigerman@umontreal.ca}
\and
Amit H. Bermano\\
Tel Aviv University, Israel\\
{\tt\small amberman@mail.tau.ac.il}
\and
Thibault Groueix\\
Adobe Research, USA\\
{\tt\small groueix@adobe.com}
}

\begin{document}

\definecolor{darkgreen}{RGB}{0,110,0}
\definecolor{darkred}{RGB}{170,0,0}
\def\greencheckmark{\textcolor{darkgreen}{\checkmark}}
\def\redxmark{\textcolor{darkred}{\text{\ding{55}}}}  %

\addeditor{amir}{AB}{0.7, 0.0, 0.7}
\addeditor{thibault}{TG}{0.0, 0.0, 0.8}
\addeditor{vova}{VK}{0.0, 0.5, 0.0}
\addeditor{amit}{AB}{0.1, 0.5, 0.9}
\addeditor{noam}{NA}{0.8, 0.0, 0.0}
\addeditor{matheus}{MG}{0.6, 0.6, 0.0}

\showeditsfalse
\showeditstrue




\twocolumn[{%
\renewcommand\twocolumn[1][]{#1}%
\maketitle
\centering
\vspace{-1.5em}
\includegraphics[width=1.0\linewidth]{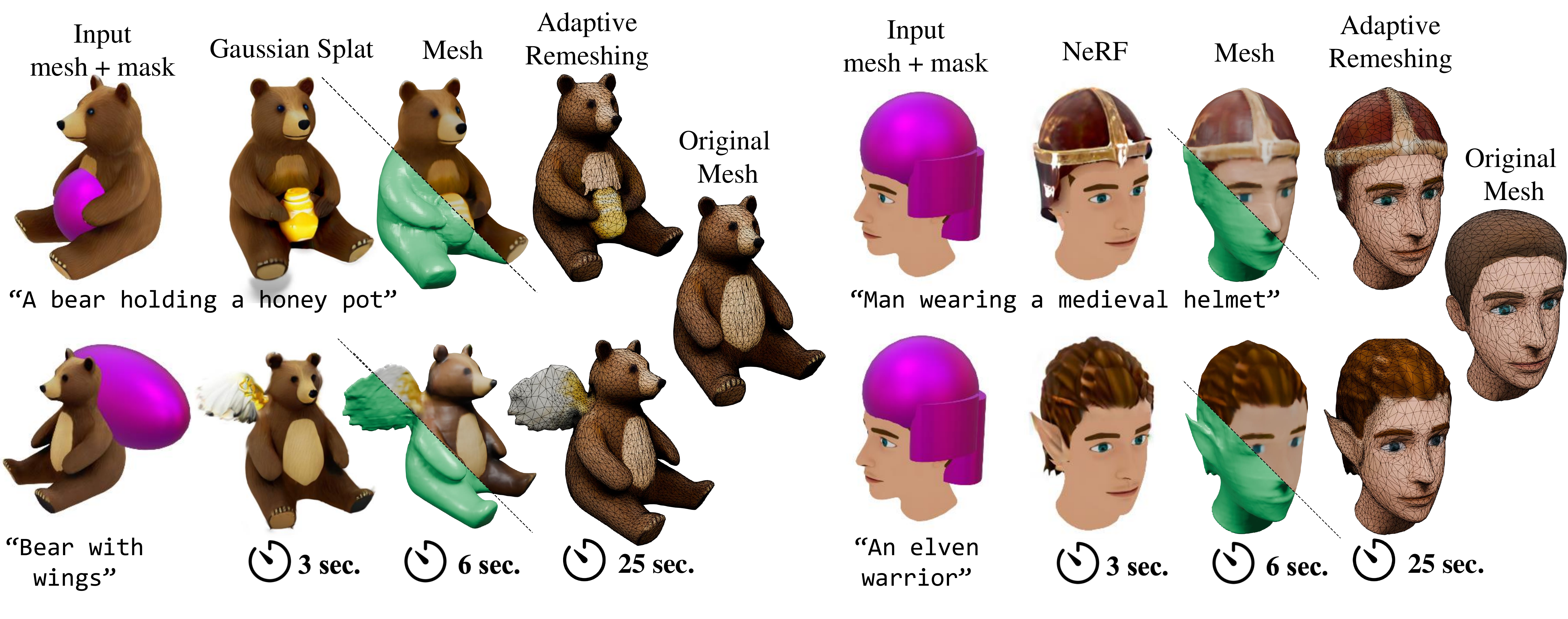}  
\captionof{figure}{Our method takes as input a 3D object along with a 3D mask (first column) and a text prompt, and uses our multiview inpainting diffusion model to consistently paint the mask in four rendered views of the object. Off-the-shelf reconstructors can be used on the multiview output to give an NeRF, a Gaussian Splat (second column), or a mesh (third column)  that can be used along with adaptive remeshing to ensure the unmasked region is exactly preserved \eg topology, uvs, (fourth and fifth column). This feedforward approach is orders of magnitude faster than previous works in generative 3D editing, \textbf{taking just $\approx3$ seconds per multiview edit}, then 0.7 seconds to reconstruct a GS or a NeRF, 3 seconds for a mesh, and $\approx20$ seconds for optional mesh post-processing.}
\label{fig:teaser}
\vspace{1em}
}]

\input{content_tex_sources/0_abstract}
\input{content_tex_sources/1_introduction}
\input{content_tex_sources/2_relatedwork}
\input{content_tex_sources/3_method}
\input{content_tex_sources/4_experiments}

\input{content_tex_sources/5_applications}

\input{content_tex_sources/6_conclusion}

{
    \small
    \bibliographystyle{ieeenat_fullname}
    \bibliography{main}
}


\end{document}

%% file: preamble.tex
\usepackage[dvipsnames]{xcolor}
\usepackage{makecell}
\usepackage{units}
\usepackage{multirow} 
\usepackage{tikz}
\usepackage{dsfont}
\usepackage{tabularray} 
\usepackage{wrapfig}
\usepackage[ruled,vlined]{algorithm2e}

\newcommand{\ourmethod}{Instant3dit\xspace}

\newcommand\ch[1][]{Type 1}
\newcommand\scribble[1][]{Type 2}
\newcommand\surface[1][]{Type 3}

\newif\ifshowedits

\newcommand{\addeditor}[3]{%
  \definecolor{#1color}{rgb}{#3}
  \expandafter\newcommand\csname #1\endcsname[1]{%
  \ifshowedits
    {\color{#1color} ##1}%
  \else
    {##1}%
  \fi
  }%
  \expandafter\newcommand\csname #1rmk\endcsname[1]{%
  \ifshowedits
    {\color{#1color} {\bf [#2: ##1]}}
  \fi
  }%
  \expandafter\newcommand\csname #1rpl\endcsname[2]{%
  \ifshowedits
    {{\color{#1color} ##1} \sout{##2}}
  \else
    {##1}
  \fi
  }%
}

\newcommand{\mycomment}[1]{}

\newcommand\customparagraph[1]{\vspace{0.7em}\noindent\textbf{#1}}

\newcommand{\bgcolor}[2]{\setlength{\fboxsep}{0pt}\colorbox{#1}{\strut #2}}
\definecolor{cbad}{HTML}{EAC2C2}
\definecolor{cmeh}{HTML}{FFE1C9}
\definecolor{cok}{HTML}{FDF3D0}
\definecolor{cgood}{HTML}{B3D09F}

\newcommand{\coloredCell}[3]{\definecolor{mycolor}{HTML}{#2}\tikz[baseline=(char.base)]{\node[fill=mycolor,inner ysep=2pt, inner xsep=5pt, minimum width=#1, text width=#1, align=right] (char) {#3};}}

\newcommand{\cellCD}[3][24pt]{\coloredCell{#1}{#2}{#3}}

\newcommand{\BGcolor}[3][HTML]{\definecolor{mycolor}{HTML}{#2}\bgcolor{mycolor}{#3}}

\newcommand{\R}{\ensuremath{\mathbb{R}}}

%% file: content_tex_sources/0_abstract.tex
\begin{abstract}
\vspace{-0.5cm}
We propose a generative technique to edit 3D shapes, represented as meshes, NeRFs, or Gaussian Splats, in $\approx3$ seconds, without the need for running an SDS type of optimization.
Our key insight is to cast 3D editing as a multiview image inpainting problem, as this representation is generic and can be mapped back to any 3D representation using the bank of available Large Reconstruction Models. We explore different fine-tuning strategies to obtain both multiview generation and inpainting capabilities within the same diffusion model. In particular, the design of the inpainting mask is an important factor of training an inpainting model, and we propose several masking strategies to mimic the types of edits a user would perform on a 3D shape. Our approach takes 3D generative editing from hours to seconds and produces higher-quality results compared to previous works. project page: 
\url{https://amirbarda.github.io/Instant3dit.github.io/} 
\end{abstract}

%% file: content_tex_sources/1_introduction.tex
\vspace*{-0.25cm}
\section{Introduction}
\label{sec:intro}
\vspace*{-0.15cm}

Recent years have seen an explosion in AI-guided 3D generative techniques for creation and editing of 3D content, allowing the user to control the generation process with often  as little as a single text prompt~\cite{sella2023voxe, weber2023nerfiller, yi2024gaussiandreamerfastgenerationtext, cheng2024progressive3dprogressivelylocalediting, chen2023fantasia3d, barda2024magicclaysculptingmeshesgenerative}. While text prompts provide an easy interface for novice users, they  lack \textit{fine control} over the generation, such as generating a specific object in a specific location over a pre-existing 3D model, e.g., make a bear hold a previously non-existent honey pot (see Figure~\ref{fig:teaser}). 

Thus, several recent works have focused on such \textit{localized} generation of 3D objects, namely 3D \emph{inpainting}~\cite{weber2023nerfiller,cao2024mvinpainter,sella2023voxe}, i.e., filling-in masked-out content in a 3D object, conditioned on a textual description of the desired fill-in, similarly to 2D inpainting~\cite{mvedit2024, weber2023nerfiller}. Figure~\ref{fig:teaser} shows several examples of such inpaintings produced by our method, where the mask (made up of coarse 3D primitives) is highlighted in purple color, the text prompt describing the target inpainting is written below, and the resulting inpainted 3D reconstructions are shown next to it. 


Current 3D inpainting methods cannot be integrated into production pipelines, as they suffer from two fundamental issues: 1) long runtimes, and 2) low quality. We observe these two issues can be explained by previous works' reliance on optimization of the 3D model by \emph{distilling} knowledge from a  generative model for 2D images via some variant of Score Distillation Sampling (SDS)~\cite{poole2022dreamfusiontextto3dusing2d}.
Such an optimization process is extremely slow, as it relies on running an image diffusion model over multiple renderings of the 3D object and back-propagating gradients, thus explaining the first issue of slow runtimes. However, we argue that it also lies at the root of the second issue, of degradation in quality: as we show through extensive comparisons (see Figure~\ref{fig:nerf-editing}), this class of approaches significantly harms the \emph{quality} of the inpainting, as SDS-like optimization produces inaccurate and fuzzy results.
We attribute it to the fact observed in the original DreamFision paper~\cite{poole2022dreamfusiontextto3dusing2d}, which stated that \textit{``2D image samples produced using SDS tend to lack diversity''} and \textit{``3D results exhibit few differences across random seeds,''} explaining this behavior with the tendency to seek specific distribution modes regardless of the seed. Thus, it stands to reason it will struggle to inpaint a region of an existing object that is not one of those seeked modes.

In this paper, we propose an alternative approach to 3D inpainting that significantly improves the quality of the inpainted results and the runtime of the process.
Drawing inspiration from previous works for fast 3D generation~\cite{li2023instant3dfasttextto3dsparseview, gslrm2024, wu2024unique3d, lu2024direct2}, we eliminate both the above issues, by turning the problem on its head:  instead of directly optimizing a 3D object using, e.g., SDS, we instead train an image generator to create 2D images of the inpainted 3D object from canonical viewpoints, and then reconstruct the newly-inpainted 3D object in a post-process, via either a feed-forward prediction~\cite{xu2024instantmeshefficient3dmesh, gslrm2024, li2023instant3dfasttextto3dsparseview}, or lightweight optimization~\cite{lu2024direct2, wu2024unique3d}. By doing so we show we avoid both the slow runtimes as well as issues with masking that arise from approaches such as SDS when applied to 3D inpainting.

Adapting this multiview generation technique to the inpainting problem leads to a core challenge: how to ensure that the inpainting of masked region in the different 2D generated image is 3D-consistent from different viewpoints (otherwise, it does not represent a coherent 3D object that can be reconstructed), while at the same time adhering to the 3D mask painted over the original 3D object. Thus, following our observation above, our main technical contribution is designing a scheme to obtain a diffusion model that provides multiview-consistent inpainting. 

In order to achieve a multiview-consistent inpainting diffuser, we devise a custom training strategy, along with a novel dataset of 3D masks for 3D inpainting.  Namely, we produce multiview consistent masks, avoiding problems resulting from occlusion (that our experiments validate is critical for performance).  We  design the dataset to support three designated editing modes with different levels of granularity (see Section~\ref{subsec:method:diffusion-inpainting})

Finally, we craft a custom training strategy to leverage the priors learned by pretrained text-conditioned image generators. 
We propose to start from a pretrained image inpainter, and use our dataset and training regime to make it multiview consistent.
We provide extensive empirical evidence to the specific design choices of our approach, e.g., compare our strategy with the opposite route -- fine-tune a multiview diffuser to perform inpainting.


We show through extensive experiments that our approach enables performing elaborate 3D editing using simple masking of a 3D region, rendering the object and the mask from multiple views, running the inpainting model, and using fast multiview reconstruction techniques to obtain the edited 3D object (see Figure~\ref{fig:teaser}). Our approach is thus agnostic to the underlying 3D representation, and we show it supports meshes~\cite{xu2024instantmeshefficient3dmesh}, Gaussian Splats~\cite{gslrm2024} and Radiance Fields~\cite{xu2024instantmeshefficient3dmesh}. 
Our experiments validate the speed of our multiview image-based representation and the higher quality of the resulting inpainting compared to previous techniques.
%
%
%
To summarize, our contributions are:
\begin{itemize}
\item The first high-quality, fast method for 3D inpainting, enabling fast and localized generation on NeRFs, Gaussian Splats and meshes.
\item A 3D masking approach along with a novel training dataset of masks for multiview inpainting, with three types of masks corresponding to different edit modes.
\item A finetuning strategy to efficiently leverage pretrained representations, along with an empirical study of alternative training strategies for multiview inpainting.
\end{itemize}

    \begin{figure*}[ht!]
    \centering
    \includegraphics[width=\linewidth]{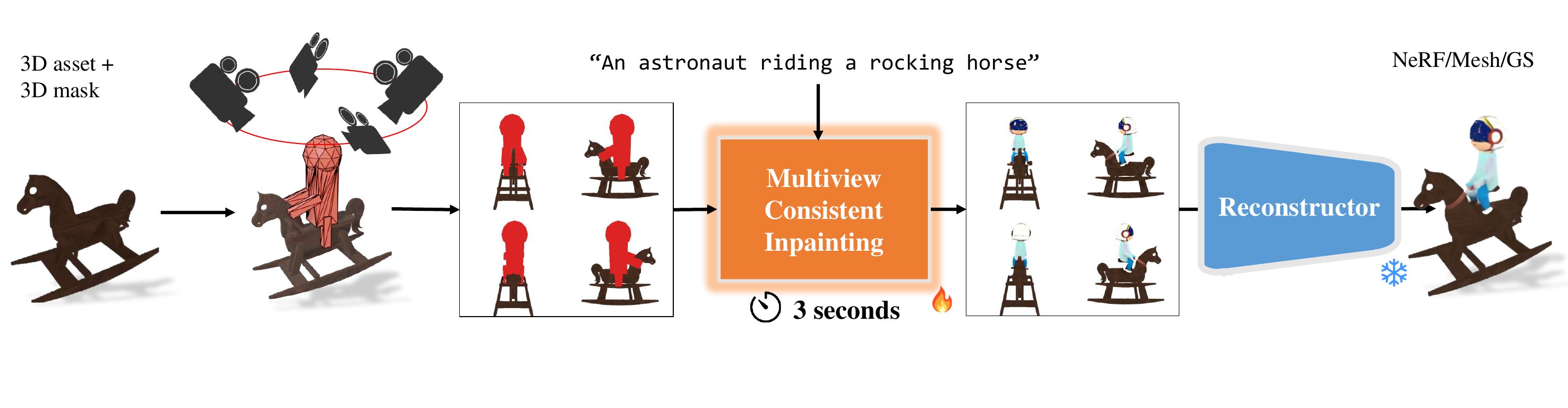}
    \caption{\textbf{Overview.} Given a NeRF, a Gaussian Splat, or a mesh, the user draws a 3D mask to mark a region to be filled and provides a text prompt to guide the generation. \ourmethod{} renders four canonical views of the masked object and uses our multiview inpainting network to fill the mask. We use off-the-shelf 3D reconstructors to convert the multiview representation into a NeRF, a Gaussian Splat, or a mesh.}
    \label{fig:enter-label}
    \vspace*{-0.15cm}
\end{figure*}


%% file: content_tex_sources/2_relatedwork.tex
\section{Related Work}
We position our work against other generative 3D editing approaches and separately discuss multiview 3D generation and inpainting. Closest to our approach are NeRFiller~\cite{weber2023nerfiller} and the concurrent work MVInpainter~\cite{cao2024mvinpainter}.

\customparagraph{3D Generative Editing.} As we aim for a general tool, we focus the discussion on category-agnostic approaches. 
The literature can be organized by the core technique used to distill the 2D generative prior - CLIP optimization~\cite{Gao_2023_SIGGRAPH, khalid2022clipmesh, michel2022text2mesh}, Score-Distillation Sampling (SDS)~\cite{kim2024meshupmultitargetmeshdeformation, barda2024magicclaysculptingmeshesgenerative, cheng2024progressive3dprogressivelylocalediting, li2023focaldreamertextdriven3dediting, chen2023fantasia3d, zhuang2023dreameditor, mikaeili2023sked, sella2023voxe, metzer2022latent, mvedit2024},  Iterative Dataset Update (IDU)~\cite{instructnerf2023, weber2023nerfiller} - or by the type of representation targeted - meshes~\cite{Gao_2023_SIGGRAPH, kim2024meshupmultitargetmeshdeformation, barda2024magicclaysculptingmeshesgenerative, michel2022text2mesh, khalid2022clipmesh}, NeRFs~\cite{mikaeili2023sked, sella2023voxe, metzer2022latent, mvedit2024, instructnerf2023, weber2023nerfiller}, or SDF~\cite{cheng2024progressive3dprogressivelylocalediting, zhuang2023dreameditor, chen2023fantasia3d, li2023focaldreamertextdriven3dediting}. All generative editing approaches take advantage of Text-to-Image (T2I) models, whose prowess stems from the billions of text-image pairs available as training data. This abundance of data does not extend to the 3d domain, with the largest public datasets, Objaverse and Objaverse-XL \cite{deitke2022objaverseuniverseannotated3d, objaverseXL}, containing millions of meshes, with the vast majority being very low-quality data.

\textit{The Score Distilation Sampling (SDS)} loss,  first defined in \cite{poole2022dreamfusiontextto3dusing2d}, use T2I models as guidance for 3d generation, by iteratively denoising renderings. SDS has multiple downsides : the denoised images lack consistency at different iterations, and the optimization is therefore very noisy, requiring careful parameter tuning to converge. Second, the optimization  typically takes around one hour to converge, as it requires a lot of steps~\cite{barda2024magicclaysculptingmeshesgenerative}. Combining SDS with Gaussian Splatting partially alleviate this problem by making the rendering operation in each  step much faster, as shown in GaussianDreamer~\cite{yi2024gaussiandreamerfastgenerationtext}, but is still not interactive (about 15 minutes per generation).  Lastly, SDS produces saturated, low-quality textures due to the high classifier-free guidance used, even though recent variations alleviate this issue~\cite{lukoianov2024score}.

\textit{Iterative Dataset Update}, first introduced in~\cite{instructnerf2023}, is a variant of SDS,  taking multiple denoising steps to produce a clean image, used for several optimization steps in a row. 
NeRFiller~\cite{weber2023nerfiller} extends this approach for editing of NeRFs.

Compared to prior work, our approach does not require an optimization to leverage the generative priorhttps://git.corp.adobe.com/pages/adobe-research/mvgenfill-client/. We propose to perform the edit using a multiview image grid to represent the 3D asset, with a single diffusion inference.  As a result, editing is orders of magnitude faster (\ie from hours to seconds) and more stable (\ie our approach always robustly produces a result and does not require careful tuning of hyperparameters). Lastly, several off-the-shelf LRM transformers~\cite{honglrm} can instantly convert our inpainted image grid to meshes~\cite{wei2024meshlrm}, NeRFs~\cite{li2023instant3dfasttextto3dsparseview}, or Gaussian Splats~\cite{gslrm2024}, making our editing technique representation agnostic.

\customparagraph{Multiview generative 3D.} 
To encourage multiview 3D consistency, several works have explored modifying the attention layers with \textit{cross}-attention blocks to facilitate the exchange of information between views~\cite{long2024wonder3d, liu2023syncdreamer, shi2024mvdreammultiviewdiffusion3d}. A simpler approach consists of directly generating a 2x2 image grid, without modifying the attention layers. In this case, the exchange of information between view is done via \textit{self}-attention. This effectively trades resolution with consistency, and can be achieved by conditionning a T2I model on multiview depth~\cite{ceylan2024matatlas, textmesh}, or by fine-tuning on a dataset of 3D multiview renderings~\cite{li2023instant3dfasttextto3dsparseview, shi2023zero123++, wang2023imagedream}. Unique3D~\cite{wu2024unique3d} and direct 2.5d~\cite{lu2024direct2} extend this approach to image grids of normals, and CRM~\cite{wang2024crm} to canonical coordinate map. 
These works have shown that adopting this representation to generate 3D shape is orders of magnitude faster than SDS, from hours to seconds. In this work, we extend these ideas to 3D editing, and explore fine-tuning strategies to get inpainting and multiview generation in the same model. Of note, NeRFiller~\cite{weber2023nerfiller} proposes to inpaint a grid of 2x2 images with a single-view inpainting model \textit{without} fine-tuning it for multiview consistency. We compare against this baseline and show the importance of fine-tuning the inpainting model. Furthermore, NeRFiller relies on a more costly IDU optimization,  which requires about 30k steps.

\customparagraph{Image inpainting.} The ability to collect and annotate images at scale has fueled major progress in image inpainting. Current inpainting tools have a wide range of applications ranging from real photographs to illustrative drawing~\cite{photoshop, stablediffusion}. We refer the reader to the recent survey of Xiang~\etal~\cite{inpaintingsurvey}. In this paper, we are interested in leveraging these pre-trained priors to edit 3D assets. As discussed in the previous paragraph, the main challenge is achieving multiview consistent results, as open-source models like Stable Diffusion~\cite{stablediffusion} inpaint each frame independently without 3D consistency. In this work, we propose to teach multiview consistency to a single-view inpainting network. As noted in Zeng~\etal~\cite{highresInpainting}, the design of the inpainting masks is a critical part of training an inpainting model. The masks should resemble the type of masks that users will draw at inference as closely as possible.  However, how to apply that insight to 3D editing is not straightforward. We propose three masking strategies that address different workflows in Section~\ref{subsec:method:diffusion-inpainting}.
%

\textit{Concurrent work.} Similarly observing that IDU and SDS are unstable and lengthy optimizations, the recent approach MVInpainter~\cite{cao2024mvinpainter} adds a video priors to a single-view inpainting model, via a LoRA~\cite{hulora}, to encourage multiview consistency. Their focus in on object insertion and removal in captured 3D scenes, while ours in on 3D generative editing of objects, which leads us to different masking strategies to train the inpainting model.

\label{sec:related_work}

%% file: content_tex_sources/3_method.tex
\section{Method}
\label{sec:method}

The input to our system consists of tuples $\langle S, M, y \rangle$ representing a 3D shape $S$, a 3D mask region $M$, and a text
description $y$ -- henceforth referred to as \emph{prompt}.
The goal of our method is to create a new shape $S^\prime$ whose areas covered by $M$ are modified to follow $y$.

\begin{figure}
\includegraphics[width=\linewidth]{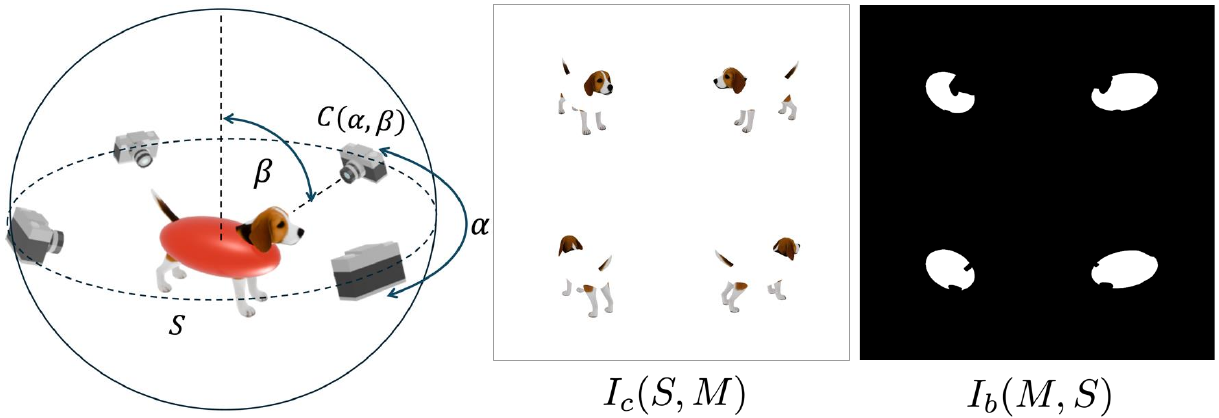}
\caption{
\label{fig:multiview-rep}
\textbf{Multiview representation.} We represent 3D shapes multiview renderings. Editing is done
using an image-based diffusion model that operates on $I_c(S, M)$ and  $I_b(M,S)$.} 
\vspace{-0.25cm}
\end{figure}

\customparagraph{Multiview representation.}
We propose to perform 3D inpainting by inpainting multiple renderings of an object in a view-consistent manner.
Consider a rendering operator $\mathcal{R}$ that renders a set of a shapes (a scene) $\mathcal{S}$, from a viewpoint $\pi$.
We refer to $\mathcal{R}_c$ and $\mathcal{R}_b$ to indicate a rendering to a \textbf{c}olor RGB image and a \textbf{b}inary mask, respectively.
Finally, we refer to $\mathcal{R}^U$ to indicate that we only render the visible pixels belonging to the shape $U$ while assuming $U \in \mathcal{S}$.
Using this operator, we define the multiview representations as :
\begin{equation}
    I_k(U, V) := \bigoplus_{\pi \in \Pi} \mathcal{R}_k^U[\{U\} \cup \{V\}; \pi] 
\end{equation}
where $\bigoplus$ concatenates the images in a $2\times2$ grid, $k$ is rendering modality (\textbf{c}olor or \textbf{b}inary),  and $\Pi = \{C(\alpha,\frac{\pi}{4}) | \alpha \in \{0, \frac{\pi}{2}, \pi, 3\frac{\pi}{2} \}\}$;
$C(\alpha, \beta)$ is a function
that returns a viewpoint configuration corresponding to a camera pointing at the origin of the coordinate system and positioned on the surface of a canonical sphere
according to azimuth $\alpha$ and elevation $\beta$.
Using this operator, the input to our multiview diffusion model is defined as $I_c(S, M)$ and $I_b(M,S)$. 
Intuitively, $I_c(S, M)$ is just an image containing the visible pixels of $S$ in the scene $\{S\} \cup \{M\}$ rendered from multiple views organized in a grid, and $I_b(M,S)$ is binary rendering of the visible pixels of $M$.
An illustration of this representation is presented in Figure~\ref{fig:multiview-rep}.


\customparagraph{3D editing.}
Given $I_c(S, M)$ and  $I_b(M,S)$, we propose to use a diffusion model $\epsilon_\theta$ to generate an inpainted multiview representation $\hat{I_c}$. From that, 
we can then use a posed multiview reconstruction process $\Phi$ to 
obtain the edited shape $\hat{S} = \Phi(\hat{I_c})$.
Different choices for $\Phi$ yield different applications and tradeoffs.
For example, we can have very fast learning-based reconstruction from posed multiview images using various representations, like
NeRFs~\cite{li2023instant3dfasttextto3dsparseview}, meshes~\cite{wei2024meshlrm} and  gaussian splats~\cite{gslrm2024}.
We can also apply lightweight optimization routines based on differentiable rendering such as ROAR~\cite{barda2023roarrobustadaptivereconstruction},
ISOMER~\cite{wu2024unique3d} or Direct2.5d~\cite{lu2024direct2}.
This class of methods is usually slightly slower but has the additional benefit of allowing us to carefully craft the optimization procedure
in order to achieve several desirable properties like geometric regularization and preservation of the original asset attributes -- color, connectivity, UVs, and so on.
Figure~\ref{fig:teaser} shows results using various types of reconstructors.
We use NeRF-LRM~\cite{li2023instant3dfasttextto3dsparseview} to compare against other NeRF editing approaches in Table~\ref{tab:training-ablation} and Figure~\ref{fig:qualitative_compaison}. 
Generating an image with $\epsilon_\theta$ takes about 3 seconds on an A100, and the various reconstructors range from a few milliseconds~\cite{li2023instant3dfasttextto3dsparseview, gslrm2024}
to a few seconds~\cite{wu2024unique3d, lu2024direct2, wei2024meshlrm}.

In the rest of this section, we explain how to train the diffusion model $\epsilon_\theta$ for multiview generation (Section ~\ref{subsec:method:diffusion}) and inpainting (Section ~\ref{subsec:method:diffusion-inpainting}). 

\subsection{Background on Diffusion Models.}
\label{subsec:method:diffusion}

\customparagraph{Training.}
During training, we sample an image $x$ from the dataset, with condition $c$ (\eg text, mask, or depth), a time step $t$ between $0$ and $T$, and a noise $\epsilon \sim \mathcal{N}(\text{0}, \text{I})$, injected to $x$ to create a noisy image $\Tilde{x}(t)$:

\begin{equation}
\Tilde{x}(t) = \sqrt{\alpha(t)} \cdot x + \sqrt{1-{\alpha(t)}} \cdot \epsilon,
\end{equation}
where $\alpha(t)$ controls the amount of noise to inject \ie $\alpha(0)=1$ is no noise and $\alpha(T)=0$ is pure noise. A denoising unet $\epsilon_\theta$ is trained to denoise $\Tilde{x}(t)$ by  minimizing:
\begin{equation}
L_\text{diff} = w(t) ||\epsilon_\theta(\Tilde{x}(t);t,c) - x ||^2, 
\end{equation}
where $w(t)$ a scheme to scale the gradients according to $t$. 
Once  $\epsilon_\theta$ is trained,  $\epsilon_\theta(\Tilde{x}(t);t,c)$ is the projection of $\Tilde{x}(t)$ to the manifold of images defined by the training dataset.
Details about the training procedure can be found in the supplemental material.

\customparagraph{Inference.} We start from pure noise $\Tilde{x}(T)$  and follow the direction of the manifold defined by $\epsilon_\theta(\Tilde{x}(t);t,c)$. There exists multiple samplers to discretize this trajectory into a discrete number of steps. In practice, we use the Euler scheduler~\cite{karras2022elucidating} and 29 steps.  

\customparagraph{Latent Diffusion.}
In this work, we use latent diffusion models \ie the diffusion happens in a 4-dimensional latent space instead of RGB space, and a pretrained VQ-VAE~\cite{Rombach_2022_CVPR} encodes and decodes images from that  space. Our method is agnostic to this, so we keep the presentation general.

\customparagraph{Multiview diffusion.}
To generate 2x2 consistent views, we follow~\cite{li2023instant3dfasttextto3dsparseview} and simply replace the training distribution with a distribution of 2x2 images. While this approach would yield poor result if it were trained from scratch, given the scarcity of high-quality 3D data, it performs well if the models are fine-tune from ``foundation'' text-to-image models, \ie pre-trained on millions of images. To create the dataset, we render a curated list~\cite{li2023instant3dfasttextto3dsparseview} of 5K objects from Objaverse~\cite{deitke2022objaverseuniverseannotated3d}, filtered for high-quality, 
and  generate high-quality captions $y_\text{n}$ for each 3D object with LLaVa~\cite{llava}. We now explain how to train for inpainting jointly.

\customparagraph{Inpainting.}
The main specificity is that the condition $c$ is composed on the text prompt $y$, the 
base image with holes, and the inpainting mask \ie $c= \{ y, I_c(S, M), I_b(M,S) \}$. In practice, for latent models,  $I_c(S, M)$ is passed through the encoder $\mathcal{E}$ of the VQ-VAE, concatenated with the downsampled version of the mask $I_b(M,S)$, and the noisy latents $\Tilde{x}(t)$, leading to a 9-channel tensor input to the denoising unet $\epsilon_\theta$, along with the encoding of the text condition. During training, we randomly drop the mask 10\% of the time, falling back to multiview diffusion training.



%

\subsection{Multi-View Inpainting Masks}
\label{subsec:method:diffusion-inpainting}
\vspace{-0.3cm}

\customparagraph{Dataset creation.}
A key part of our method consists in generating multiview masks that are 3D consistent.
Thus, the binary masks used for training our multiview inpainting model are obtained by rendering 3D shapes; \ie $I_b(M,S)$.
Consider the example in Figure~\ref{fig:multiview-rep}.
Even though $M$ is simply an ellipsoid, its multiview representation $I_b(M,S)$ is not --
it has the occlusions from its interaction with $S$.
We empirically demonstrate that 3D-aware masks and multiview consistent images are crucial for better performance (see Table~\ref{table:mask_ablation}).
Our model is trained based on a set of shapes $\mathcal{D}$.
For every shape $S\in\mathcal{D}$, we create a set of 3D masks $\mathcal{M}_S$.
Using those, we can define our training dataset $\mathcal{I}$ as:
\begin{equation}
    \mathcal{I} := \left\{ 
    \begin{array}{l}
        \big\langle I_c(S, \varnothing), I_c(S, M), I_b(M, S), y_S\big\rangle \\
        \hfill {\scriptstyle \text{where } S \in \mathcal{D}, M \in \mathcal{M}_S}
    \end{array}
    \right\}
\end{equation}
where $I_c(S, \varnothing), I_c(S, M), I_b(M, S)$ are the color ground-truth image, color input image, and binary input mask, respectively.
$y_S$ is a text prompt describing the shape $S$ obtained from a VLM model~\cite{llava}.
Since generating the masks online would slow down training, we preprocess them offline.
In practice, we have $|\mathcal{M}_S| = 30$, containing equal portions of 3 different kinds of masks.
Considering that $|\mathcal{D}| \approx 5K$, our multiview dataset $\mathcal{I}$ has $\approx150K$ data points.
We will release our dataset upon publication.

A key remaining issue is how $\mathcal{M}_S$ is created.
As noted in~\cite{highresInpainting}, the design of the training masks plays a central role in training an inpainting model.
The best performances are naturally obtained when the distribution of training masks closely follows the distribution of edits that users will make at test time.
We confirm this in our experiments with a simple ablation, where we naively use random 2D mask (see Table~\ref{table:mask_ablation}). We thus propose three types of masks, corresponding to three types of editing.
The remaining of this section will explain how each type of mask is created.

\vspace{-0.3cm}
\paragraph{Type I: coarse edit.} 
\begin{wrapfigure}[4]{r}{1.4in}
  \centering
  \vspace{-1.5em}\includegraphics[width=\linewidth]{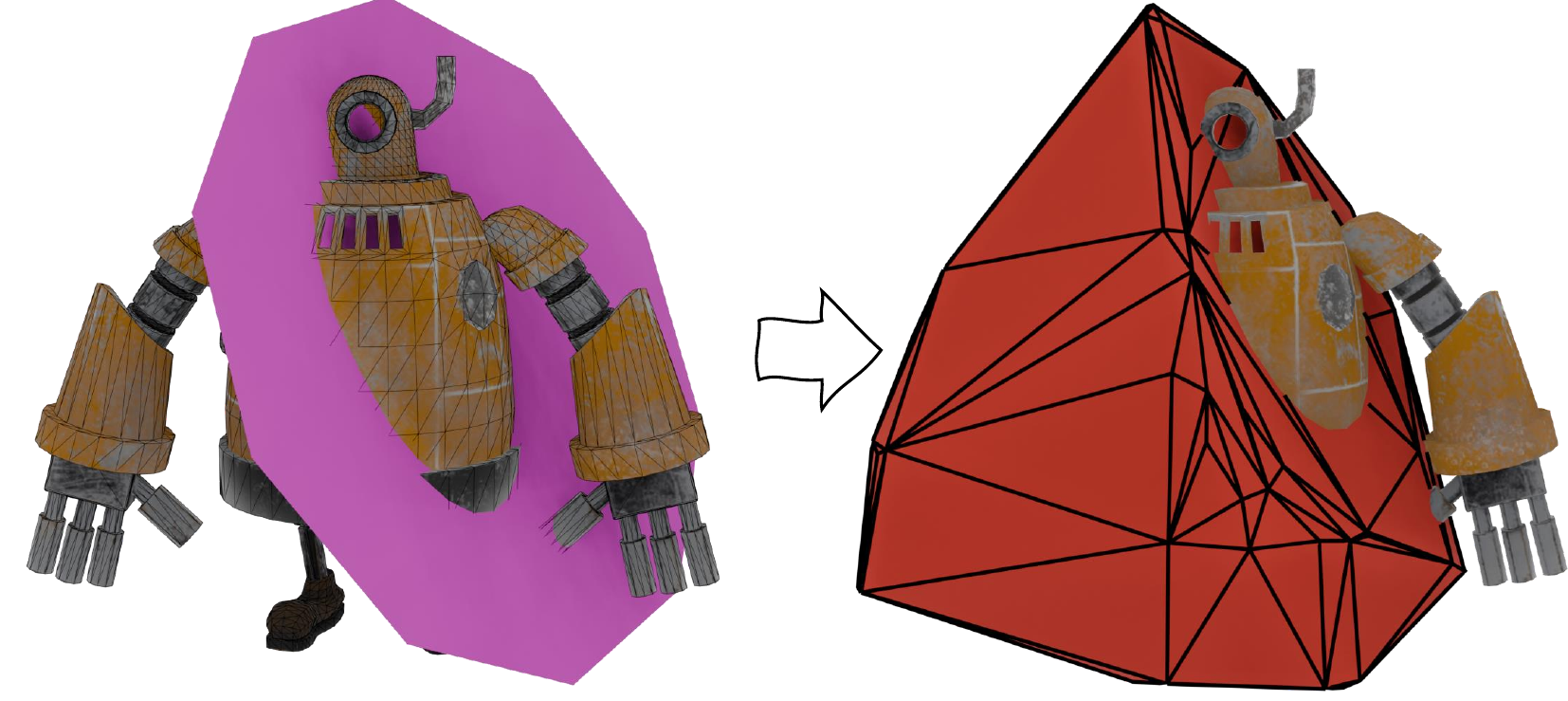}
\end{wrapfigure}
In this setting, the inpainted part of $S$ is fully contained inside $M$.
$M$ is computed by randomly sampling a part of $S$ and taking its convex hull.
To select such part, we randomly sample a plane passing through $S$, effectively splitting $S$ into two parts, and we select one part at random. 
More precisely, we start by sampling a random point $p$ inside the bounding box of $S$ and a random direction ${n}$. The plane $P$ passing through $p$ with normal $n$ is defined by $\{x \in \R^3| x\cdot p = p\cdot n\}$. $M$ is defined as the convex hull of all the face midpoints that are above $P$  \ie $\{f=(v_1, v_2, v_3)  | f \in F,  \frac{v_1 + v_2 + v_3}{3}\cdot p >= p\cdot n\}$, where $F$ denotes the list of faces. To avoid Z-fighting during rendering between $S$ and $M$, we  scale $M$ by 20\% while keeping its center of mass the same, ensuring it completely envelopes the part of $S$ 
 above $P$.

\vspace{-0.3cm}
\paragraph{Type II: mesh sculpting.}
\begin{wrapfigure}[4]{r}{1.4in}
  \centering
  \vspace{-1.5em}\includegraphics[width=1.4in]{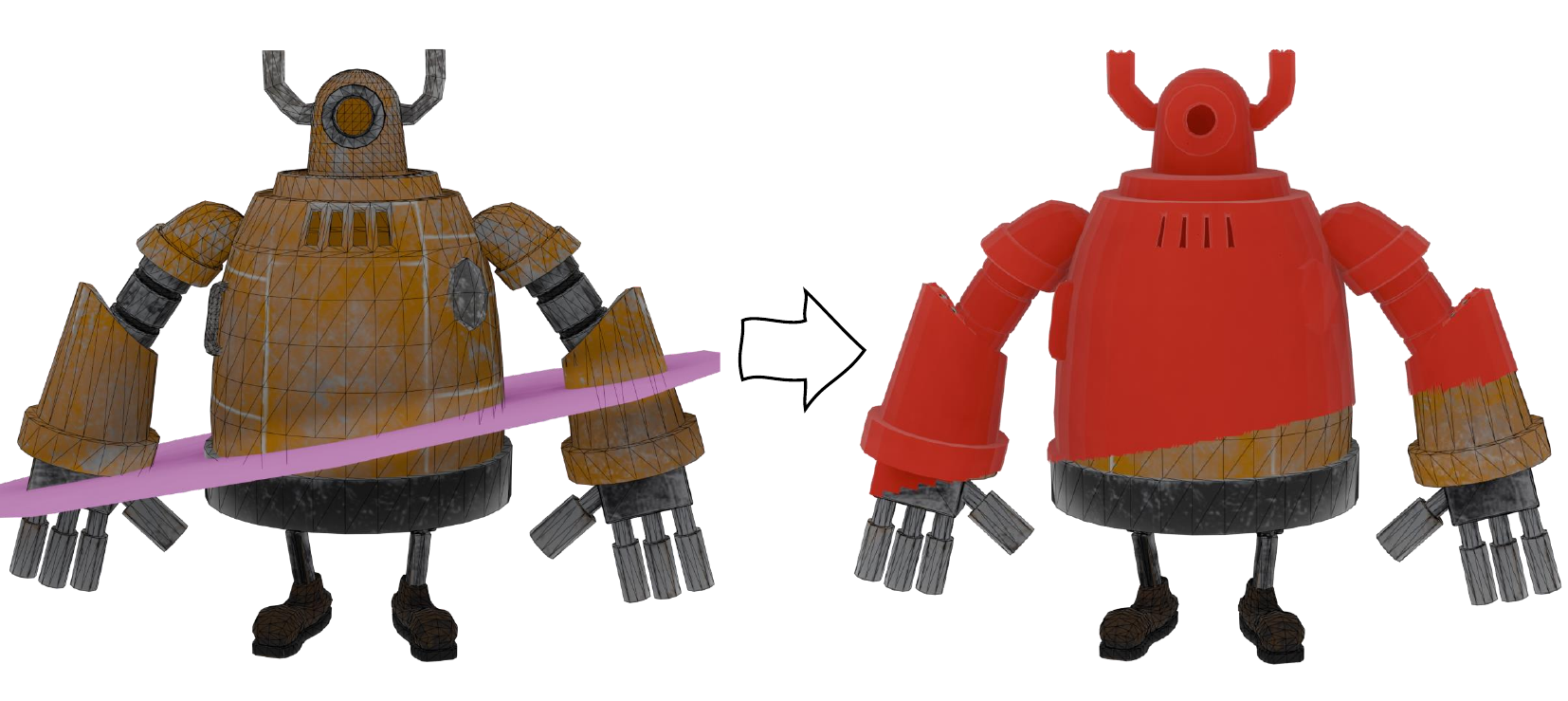}
\end{wrapfigure}
This type of mask is designed to represent a more precise edit where the user expects content to be created in a portion of space similar to the mask. 
It requires more expertise and time from the user than Type I, but also provides more precise control over the generated content.
As we can see in the illustration to the right, $M$ is a tight fit over $S$ -- no volume inside $M$ is not also inside $S$.
To generate these masks, we sample a plane $P$, as in Type I masks, and select all the faces that have their midpoint above $P$ \ie $M = \{f=(v_1, v_2, v_3)  | f \in F,  \frac{v_1 + v_2 + v_3}{3}\cdot p >= p\cdot n\}$.

\vspace{-0.3cm}
\paragraph{Type III: surface editing.}
\begin{wrapfigure}[4]{r}{1.4in}
  \centering
  \vspace{-1.5em}\includegraphics[width=\linewidth]{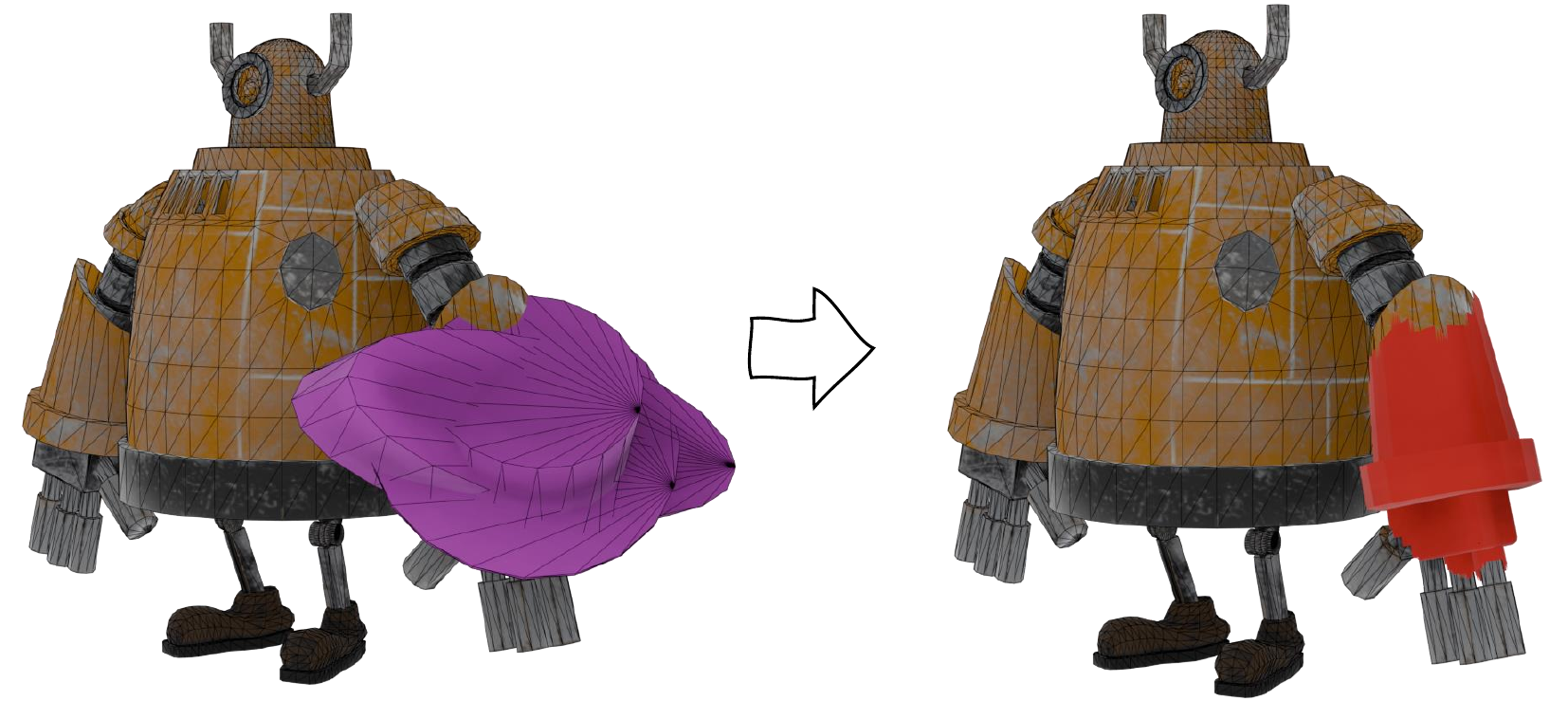}
\end{wrapfigure} We aim to support local texture modifications, where the user selects a surface patch and prompts \ourmethod{} to modify its texture.
There are various ways to generate this type of mask if we assume that the meshes have good triangulation.
Unfortunately, this is typically not the case for shape datasets.
Thus, propose a simple heuristic that works for any triangle soup.
We sample a vertex $p$ in $S$, then several cylinders with elliptical bases of varying sizes, all centered on $p$ (depicted in purple).
The number of cylinders is uniformly sampled between 3 and 6, the revolution axis is sampled on the unit sphere, the height and radii are sampled between 0.1 and 0.3.
We call this volume $C$.
Finally select all the faces whose midpoint falls within $C$ \ie $M = \{f=(v_1, v_2, v_3)  | f \in F,  \frac{v_1 + v_2 + v_3}{3}  \in C \}$ (depicted in red).

%% file: content_tex_sources/4_experiments.tex
\section{Experiments}
\label{sec:results}
\vspace*{-0.1cm}
In this section, we describe our novel multiview inpainting benchmark, which we use to compare several baselines and run ablations. \\

\vspace*{-0.1cm}
\noindent \textbf{Benchmark.}
We hold out 500 2x2 multiview images and their synthetic multiview masks from the training set to create our benchmark. We condition all inpainting on BLIP~\cite{blip} captions,
and ensure no training object is present in the benchmark. We then evaluate how well different methods inpaint all the views.
\\

\vspace*{-0.1cm}
\noindent \textbf{Evaluation Metrics.} To evaluate the quality of multiview inpainting, we use three types of metrics: measuring prompt adherence, multiview consistency, and visual quality. To measure prompt adherence, we use CLIP~\cite{clip} similarity score between the generated 2x2 multiview image and the text prompt. We follow previous work~\cite{sdxl} and use two models, CLIP-ViT-L-14 (ClipL) and CLIP-ViT-BigG-14 (ClipG), for encoding. 
To measure 3D consistency we reconstruct NeRF from the sparse inpainted views~\cite{li2023instant3dfasttextto3dsparseview}, and re-render from same camera angles as the input sparse views. We then compare the NeRF renderings to generated inpainted images using various image similarity scores: SSIM~\cite{ssim}, LPIPS~\cite{lpips} and DreamSim~\cite{dreamsim}. 
Finally, we measure visual quality using FID score~\cite{heusel2018ganstrainedtimescaleupdate} comparing to a distribution of held-out images. \\

\begin{figure}[tb]
\centering
\includegraphics[width=\linewidth]{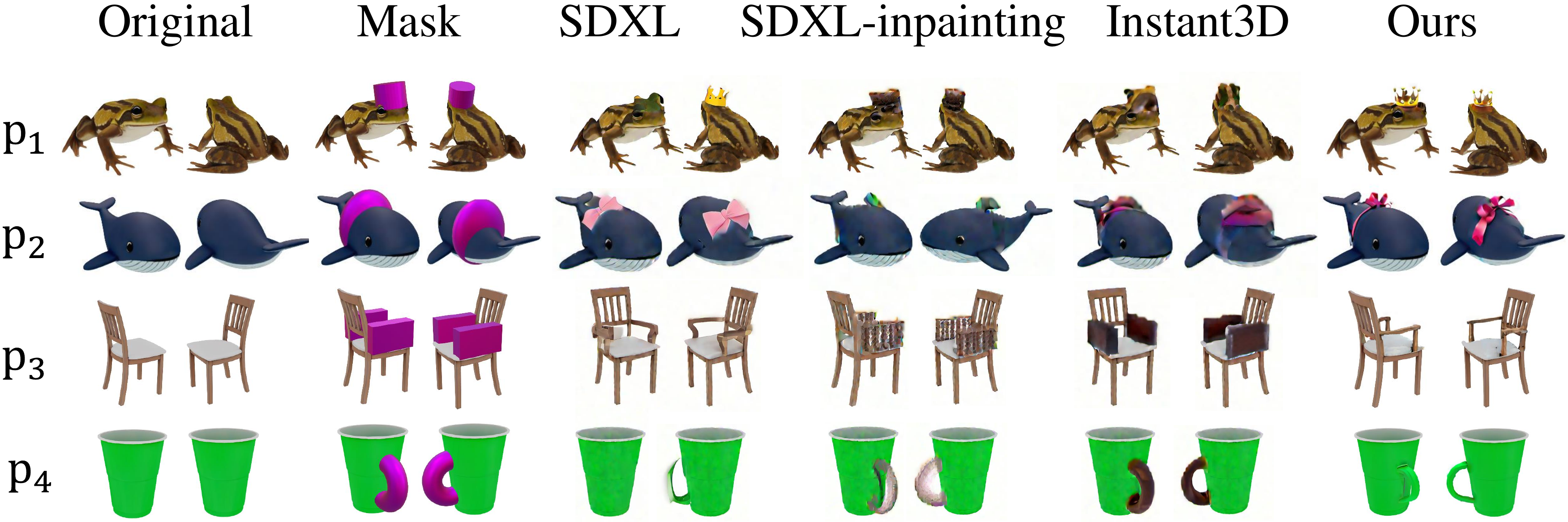}
\caption{\textbf{Comparison to baselines.} We show different inpainting results from different baselines; our multiview inpainting method offers the highest quality while maintaining consistency. 
} 
\label{fig:qualitative_compaison}
\vspace{-0.25cm}
\end{figure}

\vspace*{-0.1cm}
\noindent \textbf{Comparison to Baselines.}
We use these metrics to compare our method (Table~\ref{tab:training-ablation}, bottom) to several baseline alternatives (Table~\ref{tab:training-ablation}, top). As there are no off-the-shelf multiview consistent inpainting methods, we couple different image diffusion techniques with blended diffusion~\cite{blendeddiffusion} to inpaint the multiview image. We try two different architectures from the commonly used SDXL diffusion model~\cite{sdxl}, SDXL (as-is) and SDXL-inpainting (fine-tuned on 2D inpainting task). 
Both SDXL baselines perform poorly on multiview consistency since they are not explicitly trained to complete images consistently and, in the case of SDXL-inpainting, are only trained with 2D image masks. 
Finally we test Instant3D~\cite{li2023instant3dfasttextto3dsparseview} coupled with blended diffusion. Since this architecture is trained for 3D reconstruction it achieves a better multiview consistency, however, our method still outperforms this baseline with respect to all other metrics.
See Figure~\ref{fig:qualitative_compaison} for some example results produced with baselines and our method.
\\

\begin{table}
    \centering
    \resizebox{\linewidth}{!}{\input{tables/table_ablation_backbone}}
        \caption{\textbf{Multiview text-to-image inpainting.} This table demonstrates quantitative performance of several baselines and ablations of our method, with respect to metrics that capture prompt adherence, multiview consistency, and visual quality. 
        Results are color-coded \BGcolor{e0a4a4}{ w}\BGcolor{e7b2ac}{o}\BGcolor{eec0b5}{r}\BGcolor{f6cebe}{s}\BGcolor{fcdcc6}{t}\BGcolor{ffe4ca}{ }\BGcolor{fee8cc}{a}\BGcolor{feeccd}{n}\BGcolor{fdf0cf}{d}\BGcolor{f6f0cb}{ }\BGcolor{e3e6bd}{b}\BGcolor{cfdcae}{e}\BGcolor{bbd2a0}{s}\BGcolor{a8c992}{t }, with best highlighted in \textbf{bold}. }
    \vspace{-0.25cm}
    \label{tab:training-ablation}
\end{table}

\vspace*{-0.1cm}
\noindent \textbf{Ablating Diffusion Backbones.}
Our method can be fine-tuned from any image diffusion model, and thus we further evaluate how performance changes with respect to different backbones (Table~\ref{tab:training-ablation}, bottom section). Unsurprisingly, weaker backbones with fewer parameters (SD 1.5, SD 2.0)~\cite{Rombach_2022_CVPR} lead to inferior performance and very poor multiview consistency. Using Instant3D~\cite{li2023instant3dfasttextto3dsparseview}  gives the best multiview consistency, but fine-tuning does not seem to improve visual quality and prompt adherence. Our method fine-tunes from SDXL-inpainting~\cite{sdxl}, and generally learns to better adhere to the prompt with high visual fidelity. We believe this observation is consistent with insights from prior works~\cite{li2023instant3dfasttextto3dsparseview,weber2023nerfiller} that show that existing diffusion models already have some implicit understanding of multiview images and can be fine-tuned for 3D consistency with little training data. Inpainting, on the other hand, is a more challenging task, and thus, it is easier to fine-tune a model that was trained at scale for an inpainting task (SDXL-inpainting) to create a multiview consistent image. See supplemental material for qualitative examples. 
\\

\begin{table}[t]
\centering
    \resizebox{0.9\linewidth}{!}
 {\input{tables/mask_ablation_backbone}
 }
 \caption{\textbf{Mask ablation.} We ablate the choice of 3D masks used during training (rows), and evaluate on different subsets of the benchmark containing only some types of masks (columns).} 
\label{table:mask_ablation}
\vspace{-0.30cm}
\end{table}

\vspace*{-0.1cm}
\noindent \textbf{Ablating Masks.}

The key ingredient in our training data is the three types of multiview consistent masks. To evaluate the importance of training masks, we ablate on different types of masks, and fine-tune SDXL-inpainting model using only one mask type. 

Each row in Table~\ref{table:mask_ablation} is trained with a different type of mask: 
Random 2D follows prior 2D inpainting work~\cite{Rombach_2022_CVPR} sampling mask independently per image (see inset),

\setlength{\columnsep}{2pt}%
\begin{wrapfigure}[5]{r}{1.13in}
  \centering
  \vspace{-1.5em} \includegraphics[width=0.7\linewidth]{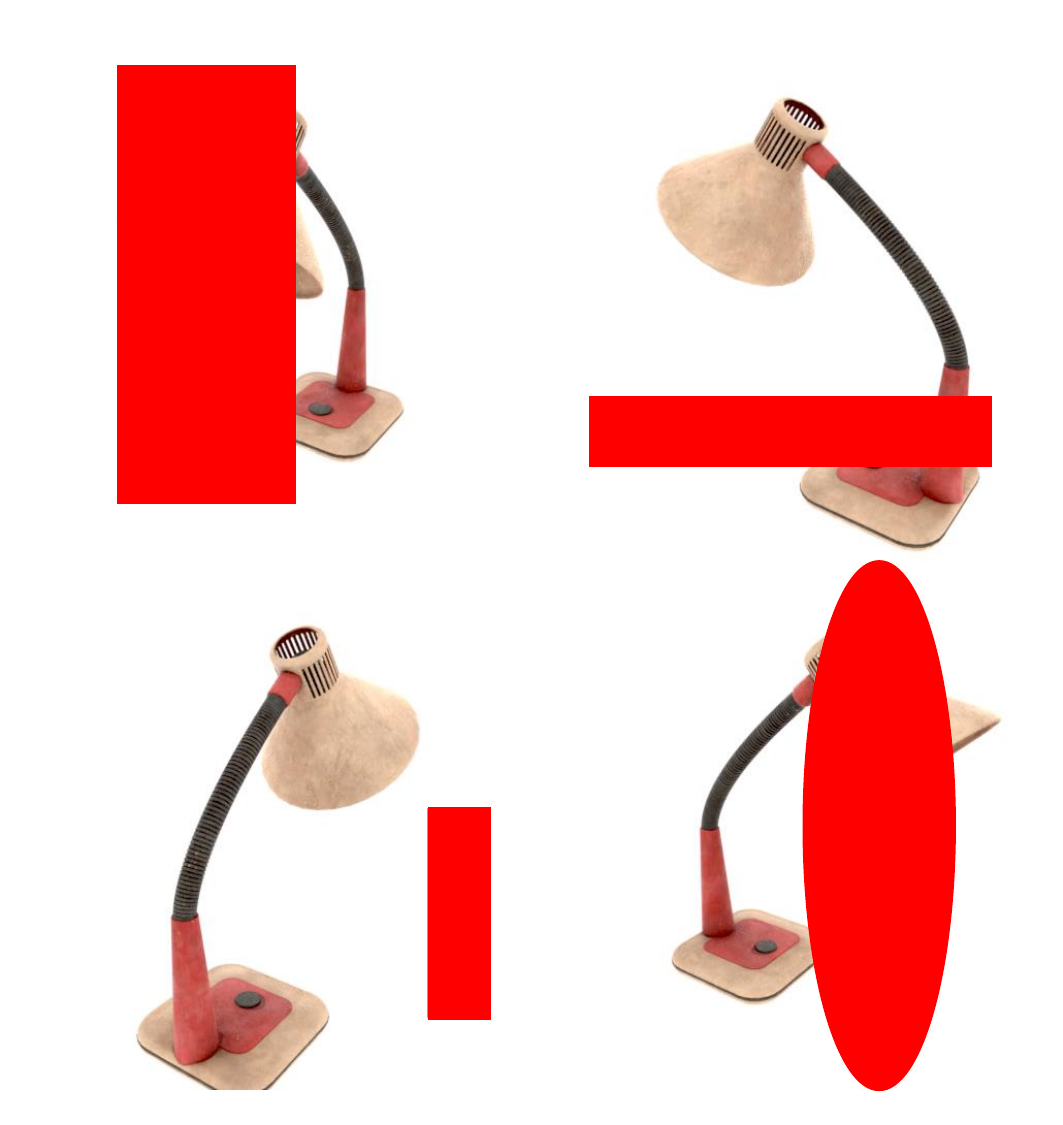}
\end{wrapfigure} 

Type I only uses coarse 3D blobs, Type II uses large surface selections, Type III uses local surface patches (see Section~\ref{subsec:method:diffusion-inpainting}). Finally, the bottom row is our method trained with all types of masks.

To compare these different techniques, we split our benchmark into single-mask vs all-masks data, and compute FID score for each test subset (see columns in Table~\ref{table:mask_ablation}). Unsurprisingly, there is a strong bias in the diagonal, indicating that training on a particular type of mask helps inpainting it at inference time. We note, however, that simple random rectangular masks perform substantially worse. Our fine-tuning on all mask types offers the best performance on arbitrary masks at inference time. 

We further take 15 user-generated masks and inpaint them to evaluate how well different types of training masks generalize to real use-cases. We found that our multi-mask training offers the highest Clip-similarity of inpainted results to the user prompts, unfortunately FID score is not meaningful for such a small sample size. We refer the reader to supplemental material where we provide qualitative inpainting results for user-generated masks, demonstrating that multi-mask training data yields the highest inpainting quality. \\



\noindent \textbf{Generalization to Novel Camera Angles.}
We note that unlike Instant3d~\cite{li2023instant3dfasttextto3dsparseview}, the inpainting model adapts its output to the orientation of the non-masked region, indicating an ability to generalize to unseen azimuth angles.
\setlength{\columnsep}{2pt}%
\begin{wrapfigure}[9]{r}{1.6in}
  \centering
  \vspace{-1.5em} \includegraphics[width=\linewidth]{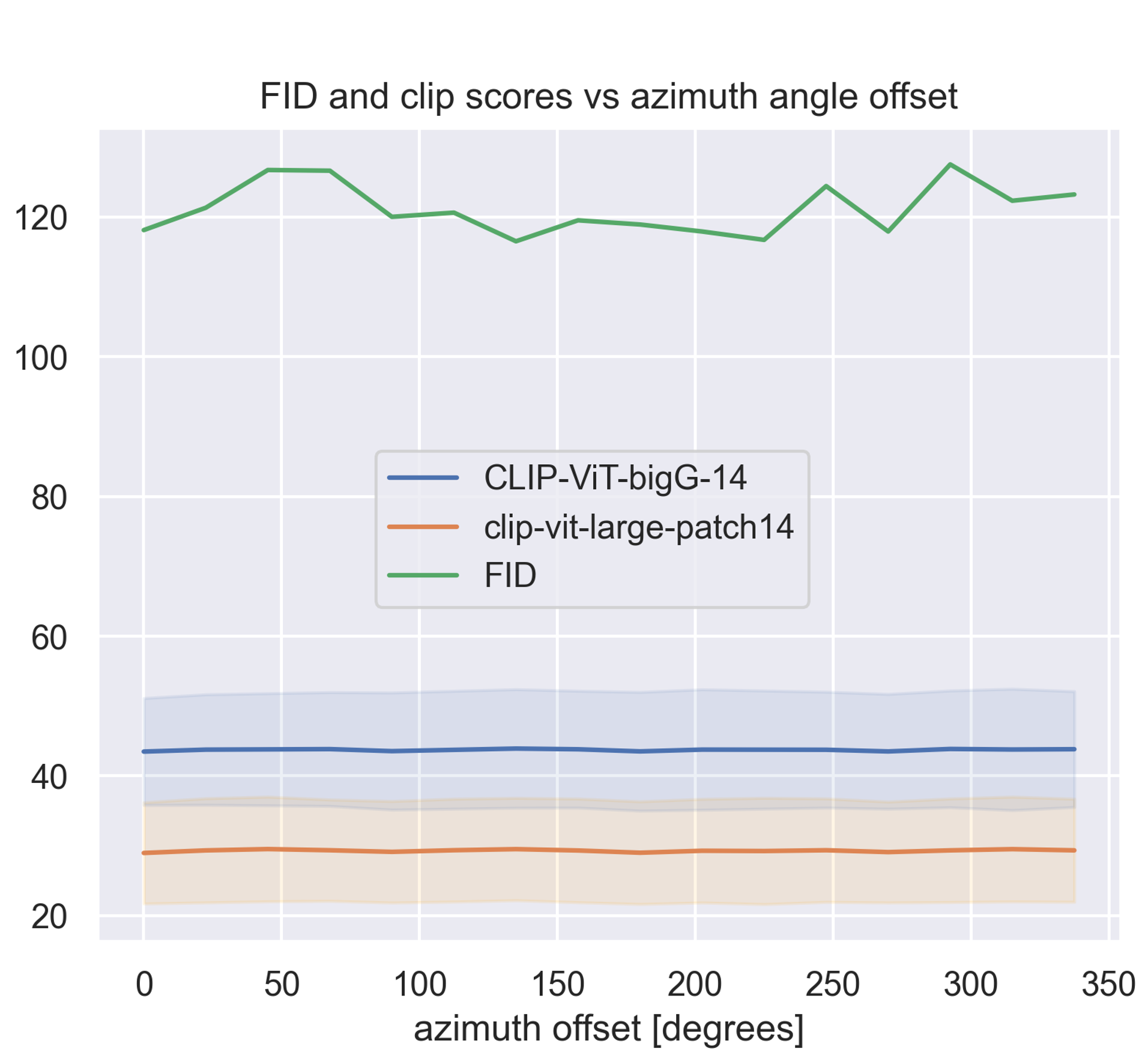}
\end{wrapfigure} 
We quantify this by offsetting the 2x2 camera positions by 16 equally spaced azimuth offsets from 0 to 337.5 degrees, generating 500 images for each offset (using the same models as in our benchmark). See how evaluation metrics change for different azimuth offsets in the inset. 

The FID, ClipL, and ClipG scores remain consistent throughout. This adaptability is crucial for inpainting, as the orientation of the unmasked portion is unknown beforehand. Fine-tuning follows \citep{li2023instant3dfasttextto3dsparseview}, with multiview renders and masks at fixed angles for each object.
Remarkably, the fine-tuned model generalizes to unseen camera orientations and FOV angles, likely due to the strong inductive bias from the masked image and prior knowledge from the baseline diffusion model. We show some visual examples in the supplementary. \\

\begin{figure}[t]
\centering
\includegraphics[width=\linewidth]{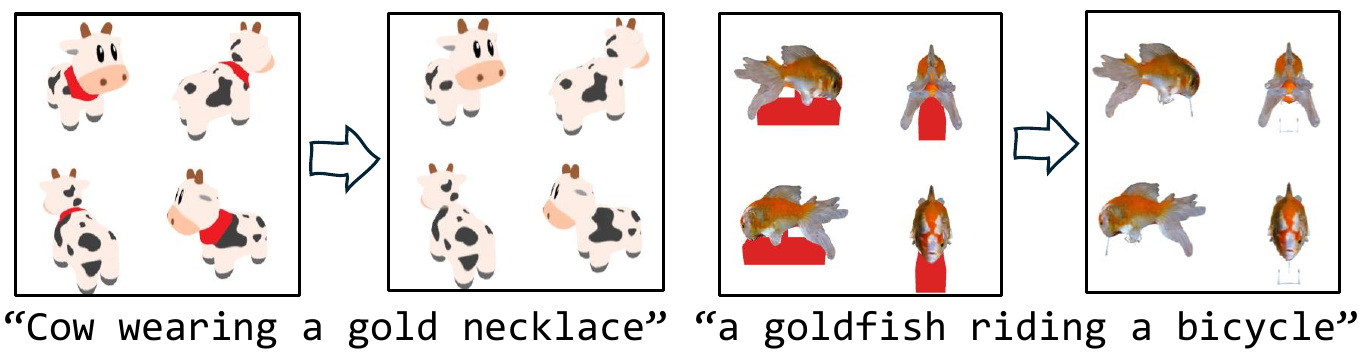}
\caption{\textbf{Failure Cases}. Typical failure cases include failure to adhere to the prompt for thin masks or large masks, which have little inductive bias from the unmasked area.}
\label{fig:limitations}
\vspace{-0.55cm}
\end{figure}

\noindent \textbf{Limitations.}
Our multiview inpainting might ignore thin masks (a behavior also observed in 2D inpainting models) (Figure~\ref{fig:limitations}, left). Additionally, due to training renders having only a white background, we find that in some cases without existing inductive bias for the mask, the network prefers to generate a white background at the expense of aligning with the prompt (Figure~\ref{fig:limitations}, right). These cases can usually be resolved by choosing a different random seed.

%% file: tables/table_ablation_backbone.tex
\setlength{\tabcolsep}{2pt}
\newdimen\wate \wate=30pt
\newdimen\wperc \wperc=24pt
\begin{tabular}{lcccccc}
\toprule
\multirow{2}{*}{Method} & \multicolumn{2}{c}{Prompt Adherence} & \multicolumn{3}{c}{Multi-view consistency} & \multicolumn{1}{c}{Visual quality} \\
\cmidrule(lr){2-3} \cmidrule(lr){4-6} \cmidrule(lr){7-7} 
 & \multicolumn{1}{c}{ClipL$\uparrow$} & \multicolumn{1}{c}{ClipG$\uparrow$} &  \multicolumn{1}{c}{SSIM$\uparrow$} & \multicolumn{1}{c}{LPIPS$\downarrow$} & \multicolumn{1}{c}{DreamSim$\downarrow$} & \multicolumn{1}{c}{FID$\downarrow$}  \\
\midrule
\midrule
SDXL~\cite{sdxl} &\cellCD[\wate]{f6f0cb}{28.63}&\cellCD[\wate]{feeccd}{42.58}&\cellCD[\wate]{ffe4ca}{0.874}&\cellCD[\wate]{e0a4a4}{0.065}&\cellCD[\wate]{e7b2ac}{0.134}&\cellCD[\wate]{cfdcae}{127.0} \\
SDXL-inpaiting~\cite{sdxl} &\cellCD[\wate]{e0a4a4}{27.57}&\cellCD[\wate]{e0a4a4}{41.33}&\cellCD[\wate]{e0a4a4}{0.857}&\cellCD[\wate]{e7b2ac}{0.064}&\cellCD[\wate]{e0a4a4}{0.137}&\cellCD[\wate]{e0a4a4}{159.5} \\
Instant3D~\cite{li2023instant3dfasttextto3dsparseview} &\cellCD[\wate]{e3e6bd}{28.78}&\cellCD[\wate]{e3e6bd}{43.06}&\cellCD[\wate]{bbd2a0}{0.892}&\cellCD[\wate]{a8c992}{0.044}&\cellCD[\wate]{bbd2a0}{0.102}&\cellCD[\wate]{a8c992}{120.0} \\
\midrule
\multicolumn{7}{c}{ablations from different diffusion backbones (our final method is at the bottom)} \\
\midrule
Ours (SD 1.5-inpainting) &\cellCD[\wate]{e7b2ac}{27.79}&\cellCD[\wate]{e0a4a4}{41.33}&\cellCD[\wate]{e0a4a4}{0.719}&\cellCD[\wate]{eec0b5}{0.09}&\cellCD[\wate]{e0a4a4}{0.599}&\cellCD[\wate]{e0a4a4}{128.3} \\
Ours (SD 2.0-inpainting) &\cellCD[\wate]{e0a4a4}{27.59}&\cellCD[\wate]{e0a4a4}{41.23}&\cellCD[\wate]{e0a4a4}{0.729}&\cellCD[\wate]{e0a4a4}{0.096}&\cellCD[\wate]{e7b2ac}{0.589}&\cellCD[\wate]{ffe4ca}{124.5} \\
Ours (Instant3D) & \cellCD[\wate]{f6f0cb}{28.57}&\cellCD[\wate]{feeccd}{42.50}&\cellCD[\wate]{a8c992}{\textbf{0.894}}&\cellCD[\wate]{a8c992}{\textbf{0.043}}&\cellCD[\wate]{a8c992}{\textbf{0.097}}&\cellCD[\wate]{a8c992}{121.1} \\
\textbf{Ours} (SDXL-inpainting) &\cellCD[\wate]{a8c992}{\textbf{29.01}}&\cellCD[\wate]{a8c992}{\textbf{43.48}}&\cellCD[\wate]{a8c992}{\textbf{0.894}}&\cellCD[\wate]{bbd2a0}{0.045}&\cellCD[\wate]{a8c992}{0.100}&\cellCD[\wate]{a8c992}{\textbf{118.4}} \\
\bottomrule
\end{tabular}

%% file: tables/mask_ablation_backbone.tex

\setlength{\tabcolsep}{2pt}
\newdimen\wate \wate=30pt
\newdimen\wperc \wperc=24pt
\begin{tabular}{lcccccc}
\toprule
\multirow{2}{*}{Method} & \multicolumn{1}{c}{Type I} & \multicolumn{1}{c}{Type II} & \multicolumn{1}{c}{Type III} &
\multicolumn{1}{c}{I+II+III} &
\multicolumn{1}{c}{User Generated} &\\
\cmidrule(lr){2-5} \cmidrule(lr){6-7}
 & \multicolumn{4}{c}{FID$\downarrow$} & \multicolumn{1}{c}{ClipG$\uparrow$}  \\
\midrule
\midrule
Random 2D &\cellCD[\wate]{e3e6bd}{145.4}&\cellCD[\wate]{e3e6bd}{129.6}&\cellCD[\wate]{e0a4a4}{102.3}&\cellCD[\wate]{feeccd}{131.1}&\cellCD[\wate]{e0a4a4}{24.22}\\
Type I &\cellCD[\wate]{a8c992}{131.2}&\cellCD[\wate]{e3e6bd}{128.6}&\cellCD[\wate]{e0a4a4}{102.3}&\cellCD[\wate]{bbd2a0}{121.3}&\cellCD[\wate]{cfdcae}{26.24}\\
Type II &\cellCD[\wate]{ffe4ca}{170.9}&\cellCD[\wate]{a8c992}{\textbf{122.6}}&\cellCD[\wate]{f6cebe}{101.7}&\cellCD[\wate]{fdf0cf}{128.2}&\cellCD[\wate]{feeccd}{25.53}\\
Type III &\cellCD[\wate]{e0a4a4}{193.0}&\cellCD[\wate]{e0a4a4}{148.1}&\cellCD[\wate]{a8c992}{\textbf{99.05}}&\cellCD[\wate]{e0a4a4}{142.2}&\cellCD[\wate]{e0a4a4}{24.2}\\
I+II+III &\cellCD[\wate]{a8c992}{\textbf{130.0}}&\cellCD[\wate]{bbd2a0}{124.9}&\cellCD[\wate]{e3e6bd}{100.0}&\cellCD[\wate]{a8c992}{\textbf{118.4}}&\cellCD[\wate]{a8c992}{\textbf{26.5}}\\
\bottomrule
\end{tabular}

%% file: content_tex_sources/5_applications.tex
\section{Applications}
While the main contribution of our work is the multiview consistent inpainting method, which can be trivially complemented with a large reconstruction model (LRM) to infer the full 3D shape from the completed views. In this section we explore applications of our technique for editing various neural and traditional 3D representations.
We also developed an interactive application where users can load their own 3D shapes, create masks with basic primitives and visualize the results of our method.
More information about this application can be found in supplemental. \\

\noindent\textbf{NeRF Editing.}
Given an input NeRF, user's mask and a prompt, we use our multiview inpainting to generate the edited views. These views are used as input to the  LRM from Instant3D~\cite{li2023instant3dfasttextto3dsparseview} to create the modified representation. We show a few edits in Figure~\ref{fig:enter-label} (our result is on the right). In addition to our method we show results for several alternative techniques that allow editing NeRFs. Vox-E \cite{sella2023voxe} and MVEdit \cite{mvedit2024} do not allow for a user-provided mask, and instead infer the area to be edited using the attention weights from the prompt, which may result in undesirable changes in texture and geometry throughout the object in areas the user would like to keep unchanged. Progressive3d \cite{cheng2024progressive3dprogressivelylocalediting} and NeRFFiller \cite{weber2023nerfiller} struggle with multiview consistency, as they rely on 2D diffusion models for SDS and IDU, respectively. This may result in incorrectly positioned edits, as in the case of the centaur example. 
Note that our approach is also significantly faster than all existing timelines and produces results in seconds rather than minutes.

We further conduct an informal user preference study between our method and the closest approach that can also inpaint multiview masks, NeRFiller~\cite{weber2023nerfiller}. We showed the masked input and the prompt to 15 users and asked them to choose between two outputs (see supplemental for exact verbiage). Out of 208 pairs, the users preferred our method in 180 cases (86\%) vs 28 (14\%) for NeRFiller. 

Since our multiview inpainting is not tied to any particular 3D representation, our method can  also be used to edit Gaussian Splats (GS) by simply using an appropriate LRM~\cite{gslrm2024}, see the supplementary for GS editing examples.\\

\begin{figure}[t]
\centering
\includegraphics[width=\linewidth]{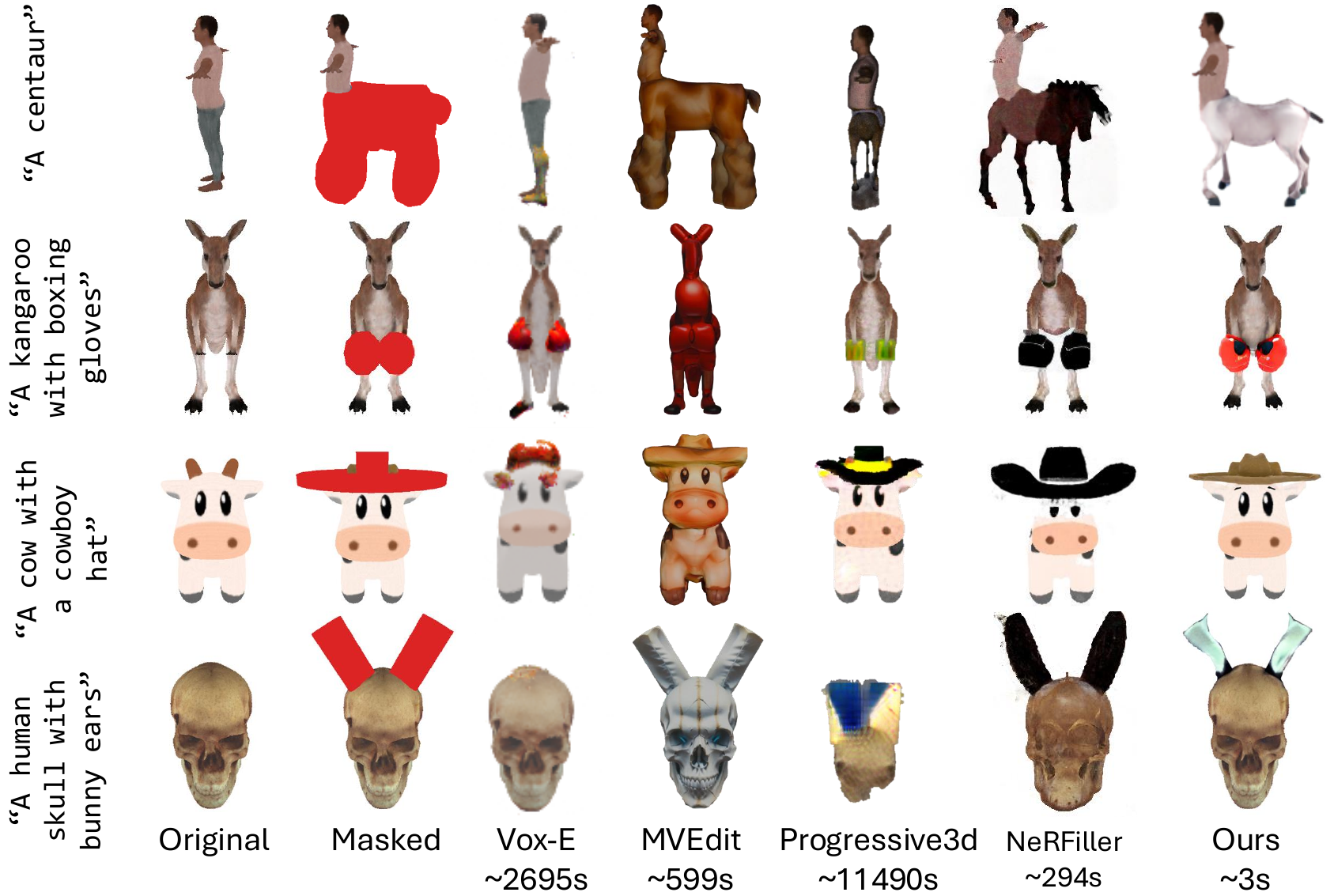}
%
%
\caption{\textbf{Application: NeRF editing.} We compare our method (rightmost column) to other methods that allow editing NeRFs, by rendering a front-facing view of the 3D edited asset for all methods. Note that the only method that takes arbitrary masks as input is NeRFiller~\cite{weber2023nerfiller}, we also conduct an informal user study with 15 users choosing between our output and NeRFiller for 208 pairs of results. We found that users prefer our method in 86\% of the cases. We report average timing for all methods, and confirm \ourmethod{} is  substantially faster than other alternatives.}
\label{fig:nerf-editing}
\vspace{-0.25cm}
\end{figure}

\begin{figure}[t]
\centering
\includegraphics[width=\linewidth]{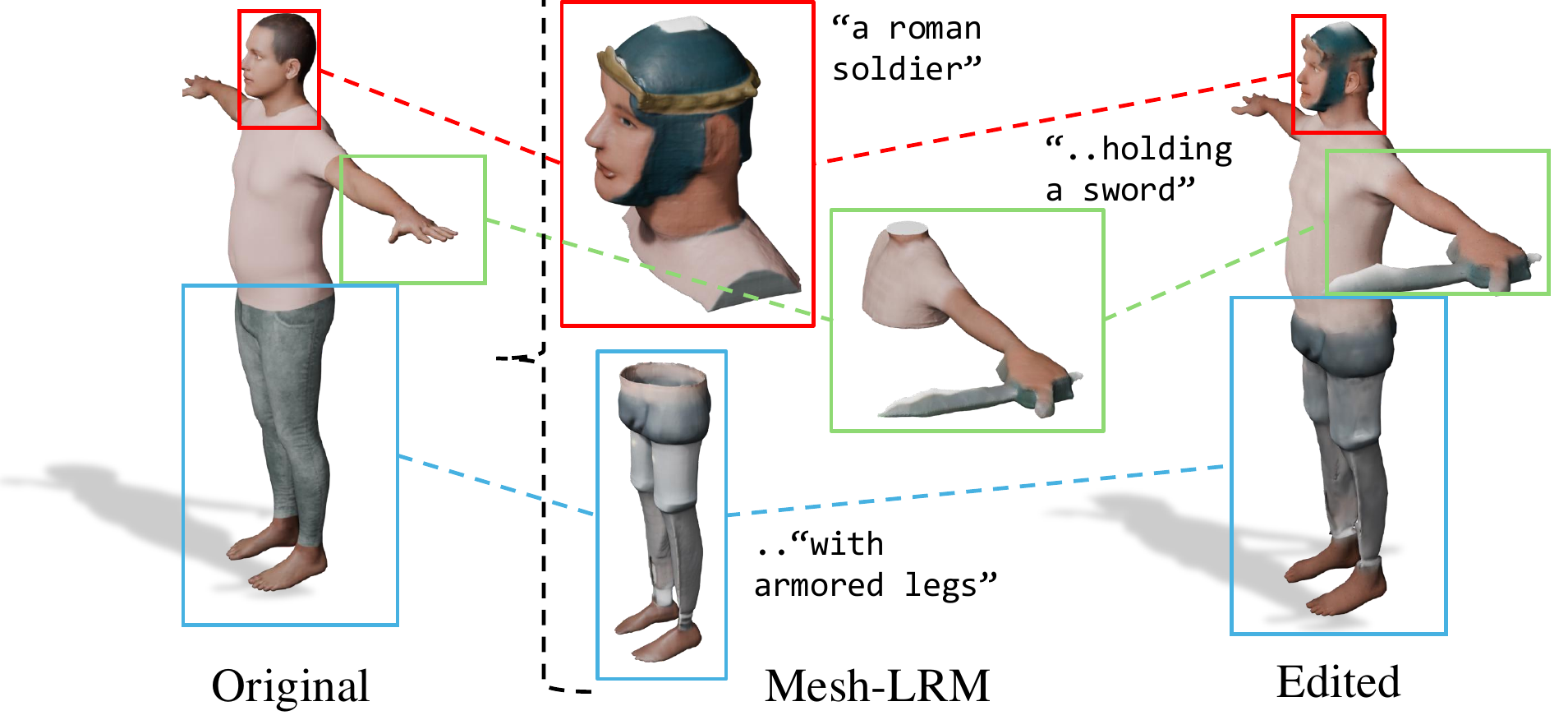}


\caption{\textbf{Application: mesh editing}. Each of the 3 edits shown is performed locally on the original mesh, only modifying regions selected by the user. Locality is achieved by running the ROAR~\cite{barda2023roarrobustadaptivereconstruction} geometry optimization, which takes ~20 extra seconds per edit.}
\label{fig:mesh_editing}
\vspace{-0.25cm}
\end{figure}


\noindent \textbf{Mesh Editing.}
For the users that work on traditional mesh representation, we propose using MeshLRM \citep{wei2024meshlrm} with an adaptive remeshing layer \cite{barda2023roarrobustadaptivereconstruction} that uses the LRM result as a guidance. This leads to a fast and fully controllable mesh editing pipeline that is guaranteed to preserve the original mesh attributes (e.g., UVs, rigging) and triangulation in the unedited parts of the mesh~\cite{barda2024magicclaysculptingmeshesgenerative}. 
We noticed that the guidance meshes tend to be over-smoothed due to the relatively low resolution of the LRM's triplane. To maintain fine details, we use a normal estimator~\cite{fu2024geowizardunleashingdiffusionpriors} directly on the diffusion output and use the normals as targets in the vertex optimization. See Figure~\ref{fig:mesh_editing}, supplemental figures, and a video for mesh editing examples.   \\

\begin{figure}[t]
\centering
\includegraphics[width=0.8\linewidth]{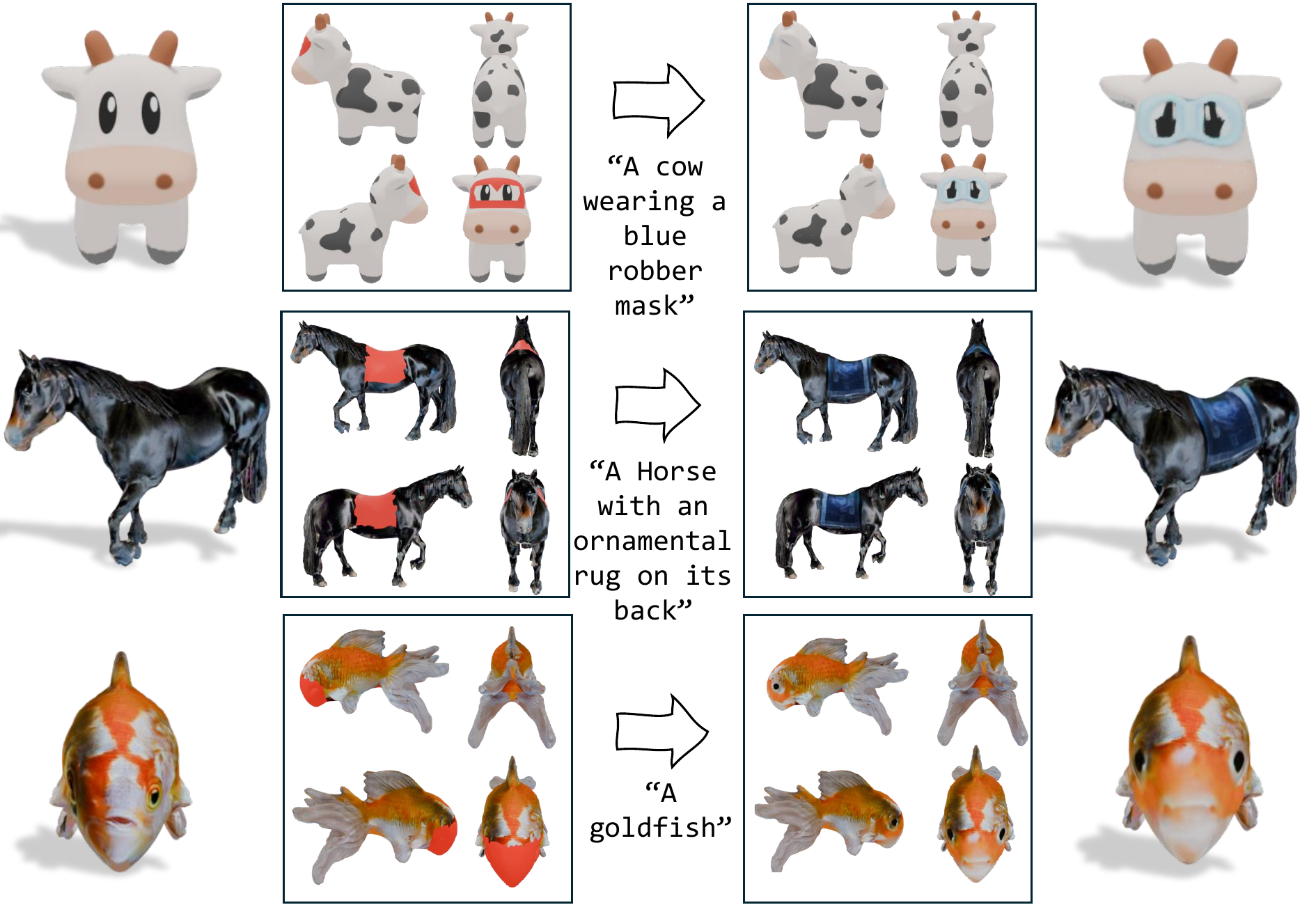}
\caption{\textbf{Application: texture editing.} Our method can be used to modify  texture on a user-selected region. In this case, we run through our NeRF editing pipeline, but only sample colors from the NeRF in the selected region. 
}
\label{fig:texturing}
\vspace{-0.25cm}
\end{figure}

\noindent \textbf{Texture Editing.}
The user can use our method to edit surface texture details. We use NeRF-LRM to reconstruct a 3D representation and back-project the colors to the user-selected mesh region. 
Starting with an auto-generated texture from previous work~\cite{richardson2023texturetextguidedtexturing3d} (Fig.~\ref{fig:texturing}, first column), the user labels a mask (second column) to either add new texture elements (facemask, saddle) or fix artifacts in the texture (goldfish exhibiting inconsistent left-right colors due to lack of multiview consistency in the previous texturing method).

%% file: content_tex_sources/6_conclusion.tex
\section{Conclusion}
We introduced \ourmethod{}, a multiview inpainting diffusion model, and show its application to \textit{fast} and \textit{localized} editing of 3D assets. We propose three types of inpainting masks to train the model, corresponding to three types of user edits, and plan to release this dataset. We justify our training strategy and choice of 3d masks by measuring a comprehensive set of metrics: prompt adherence, 3d consistency of the generated content, and generation quality. \ourmethod{} shows superior quality to the contemporary 3D editing pipelines and is orders of magnitude faster, running as fast as a single image generation using diffusion.

We see clear opportunities to improve this tool further. First, several approaches such as DMD~\cite{dmd}  distill diffusion models into one-step or few-step models, which would make \ourmethod{} run in a fraction of a second and bring a truly interactive editing experience. Second, video diffusion model~\cite{genmo2024mochi, brooks2024video} have shown remarkable 3D consistency and visual quality, which can be harnessed to improve the quality of 3D generation further, although, in terms of speed, they are currently much slower than text-to-image models. 

\label{sec:conclusion}

